\documentclass[conference]{IEEEtran}
\IEEEoverridecommandlockouts
\usepackage{cite}
\usepackage{amsmath,amssymb,amsfonts}
\usepackage{amsthm, amssymb}
\usepackage{algorithmic}
\usepackage{graphicx}
\usepackage{textcomp}
\usepackage{xcolor}
\usepackage{booktabs}
\usepackage{subfig}
\usepackage{multirow}
\usepackage{makecell}
\newtheorem{theorem}{Theorem}
\newtheorem{lemma}{Lemma}
\newtheorem{definition}{Definition}

\def\BibTeX{{\rm B\kern-.05em{\sc i\kern-.025em b}\kern-.08em
    T\kern-.1667em\lower.7ex\hbox{E}\kern-.125emX}}
\begin{document}

\title{GED-Consistent Disentanglement of Aligned and Unaligned Substructures for Graph Similarity Learning
}

\author{\IEEEauthorblockN{1\textsuperscript{st} Zhentao Zhan}
\IEEEauthorblockA{\textit{Hangzhou Dianzi University}\\
Hangzhou, China \\
zhanzhentao@hdu.edu.cn}
\and
\IEEEauthorblockN{2\textsuperscript{nd} Xiaoliang Xu}
\IEEEauthorblockA{\textit{Hangzhou Dianzi University}\\
Hangzhou, China \\
xxl@hdu.edu.cn}
\and
\IEEEauthorblockN{3\textsuperscript{rd} Jingjing Wang}
\IEEEauthorblockA{\textit{Hangzhou Dianzi University}\\
Hangzhou, China \\
 Wangj3573@163.com}
\and
\IEEEauthorblockN{4\textsuperscript{th} Junmei Wang}
\IEEEauthorblockA{\textit{Hangzhou Dianzi University}\\
Hangzhou, China \\
jmwang@hdu.edu.cn}
}

\maketitle

\begin{abstract}
Graph Similarity Computation (GSC) is a fundamental graph-related task where Graph Edit Distance (GED) serves as a prevalent metric. GED is determined by an optimal alignment between a pair of graphs that partitions each into \textit{aligned} (zero-cost) and \textit{unaligned} (cost-incurring) substructures. However, the solution for optimal alignment is intractable, motivating Graph Neural Network (GNN)-based GED approximations. Existing GNN-based GED approaches typically learn node embeddings for each graph and then aggregate pairwise node similarities to estimate the final similarity. Despite their effectiveness, we identify a fundamental mismatch between this prevalent node-centric matching paradigm and the core principles of GED. This discrepancy leads to two critical limitations: (1) a failure to capture the global structural correspondence 
for optimal alignment, and (2) a 
misattribution of edit costs by learning from spurious node-level signals.

To address these limitations, we propose \textbf{GCGSim}, a GED-consistent graph similarity learning framework that reformulates the GSC task from the perspective of graph-level matching and substructure-level edit costs.
Specifically, we make three core technical contributions.
First, we design a \textbf{G}raph-\textbf{N}ode \textbf{C}ross \textbf{M}atching (GNCM) mechanism to learn pair-aware contextualized graph representations.
Second, we introduce a principled \textbf{P}rior \textbf{S}imilarity-\textbf{G}uided \textbf{D}isentanglement (PSGD) mechanism, justified by variational inference, to unsupervisedly separate graph representations into their aligned and unaligned substructures.
\textbf{Finally, we employ} an \textbf{I}ntra-\textbf{I}nstance \textbf{R}eplicate (IIR) consistency regularization to learn a canonical representation for the aligned substructures. 
Extensive experiments on four benchmark datasets show that GCGSim achieves state-of-the-art performance. 
Our comprehensive analyses further validate that the framework successfully learns disentangled and semantically meaningful substructure representations.

\end{abstract}

\begin{IEEEkeywords}
Graph similarity computation, Graph edit distance, Optimal alignment, Disentanglement
\end{IEEEkeywords}

\section{Introduction}

\begin{figure*}[ht]
	\centering

        \subfloat[]{\includegraphics[width=0.163\linewidth]{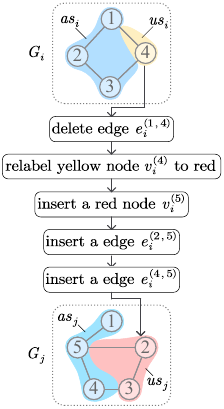}
        \label{fig:GED}}
        \hfill
        \subfloat[]{\includegraphics[width=0.787\linewidth]{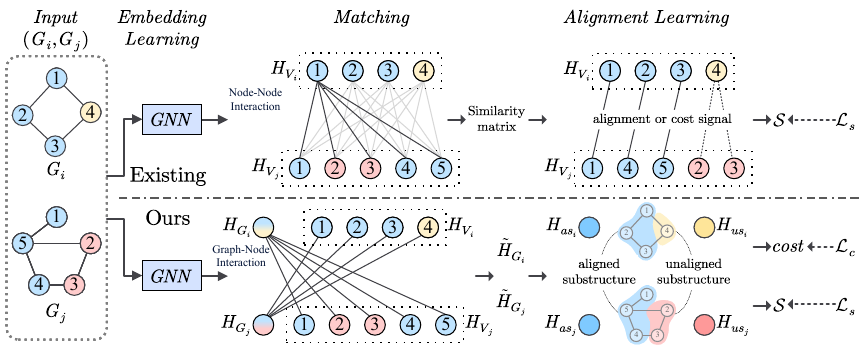}
        \label{fig:framework_comparison}}

    \caption{(a) A graph edit path for an input graph pair $(G_i, G_j)$. The graph is partitioned into aligned substructures ($as_i, as_j$) and unaligned substructures ($us_i, us_j$). GED is derived from the cost of transforming $us_i$ to $us_j$. The number within each node denotes its ID, and the color indicates its label. (b) Comparison of the conventional node-centric framework with our proposed GED-consistent framework (GCGSim).
    }
    
\end{figure*}

Graph Similarity Computation (GSC) is a fundamental task for analyzing graph-structured data.
Its central function is the querying and retrieving structurally similar entries within graph databases, a cornerstone application across numerous scientific and industrial domains such as bioinformatics, cheminformatics, and program analysis.
For example, in drug discovery, GSC enables efficient querying of compound databases to identify molecules with similar structures to a target, thereby accelerating the screening process.

A prevalent and theoretically grounded metric for GSC is the Graph Edit Distance (GED)~\cite{b1}. GED quantifies the dissimilarity between two graphs, $G_i$ and $G_j$, as the minimum cost of edit operations (node/edge insertions, deletions, and relabel) required to transform $G_i$ into $G_j$. The computation of GED can be formulated as an optimal alignment problem
, which partitions each graph into two disjoint sets: an \textbf{aligned substructure} and an \textbf{unaligned substructure}, as illustrated in Fig.~\ref{fig:GED}. 
The aligned substructures represent the structurally identical portions identified by the optimal alignment, thereby incurring zero edit cost. The final GED value is derived exclusively from the cumulative cost of editing the unaligned substructures. The NP-hard nature of exact GED computation has catalyzed the development of efficient, learning-based approximation methods.

Recently, Graph Neural Networks (GNN) have become the dominant approach for GED approximation, achieving remarkable performance~\cite{b9,b41,b24,b10,b25}. A prevalent paradigm of existing GNN-based methods follows a three-stage, end-to-end paradigm (illustrated in upper part of Fig.~\ref{fig:framework_comparison}):
(1)~\textit{Embedding Learning}: 
GNN independently encode nodes of each graph into feature vectors. 
(2)~\textit{Matching}: 
A similarity matrix is computed via pairwise node-node interactions across the two graphs, capturing node-level correspondence 
(3)~\textit{Alignment Learning}:
A final regression head maps the alignment (or cost) signals within the similarity matrix to a predicted similarity score. 
While this node-centric paradigm has proven highly effective, we identify a fundamental mismatch between such approaches and the core principles of GED.
This discrepancy leads to two critical limitations:

\underline{Limitation I:} \textit{Neglect of Global Structural Correspondence.} Existing methods, by focusing on local node-to-node matching, fail to capture the global alignment perspective 
to GED. The optimal GED alignment is a graph-level property that maximizes the isomorphism between aligned substructures. In contrast, local matching can erroneously align nodes with similar local neighborhoods but starkly different global roles (e.g., aligning a central hub in $G_i$ with a peripheral node in $G_j$).

\underline{Limitation II:} \textit{Misattribution of Edit Costs.} GED is sourced only from operations on unaligned substructures. However, existing methods learn from a dense matrix of all node-level similarities (or costs)~\cite{b43,b42}. They cannot distinguish between the "true" dissimilarity that contributes to GED (i.e., between unaligned parts) and other "irrelevant" dissimilarities (e.g., between an aligned part of $G_i$ and an unaligned part of $G_j$). This conflation leads to the model learning from spurious signals and misattributing the sources of the final graph similarity.

To address these limitations, we propose a novel framework, \textbf{GCGSim} (\textbf{G}ED \textbf{C}onsistent \textbf{G}raph \textbf{Sim}ilarity Learning), illustrated in lower part of Fig.~\ref{fig:framework_comparison}, designed to align the learning process with the core principles of GED. GCGSim reformulates the task from the perspective of graph-level matching and substructure-level edit costs. Our work addresses three core challenges:

\underline{Challenge I:} \textit{How to learn a graph representation that is aware of the alignment context?} To overcome the locality of node-level matching, we must 
learn graph embeddings that are contextualized for a specific graph pair. We propose \textbf{GNCM} (\textbf{G}raph-\textbf{N}ode \textbf{C}ross \textbf{M}atching), a cross-attention mechanism that computes a pair-aware embedding for each graph. It explicitly models the alignment between the local structures of one graph and the global structure of the other, allowing the model to dynamically highlight structurally congruent substructures.

\underline{Challenge II:} \textit{How to disentangle representations for aligned and unaligned substructures without direct supervision?} The aligned/unaligned partition is a latent property of the optimal alignment and is not available as a label during training. To solve this, we introduce \textbf{PSGD} (\textbf{P}rior \textbf{S}imilarity-\textbf{G}uided \textbf{D}isentanglement). PSGD is a principled mechanism that disentangles the pair-aware graph embedding into distinct representations for 
the aligned and unaligned substructures. Crucially, we provide a theoretical justification for PSGD from a \textit{variational inference} perspective, 
showing that it uses the model's own graph similarity estimate as a dynamic
signal to guide the disentanglement process.

\underline{Challenge III:} \textit{How to learn a canonical representation for aligned substructures and predict GED-consistent costs?}
Following disentanglement, the resulting aligned embeddings are not guaranteed to be semantically identical, as they originate from different graphs.
To resolve this, we introduce \textbf{IIR} (\textbf{I}ntra-\textbf{I}nstance \textbf{R}eplicate), a novel consistency regularization method that encourages the aligned embeddings to learn into a canonical representation. 
With this canonical representation established, our framework then interacts the substructure pairs to learn the edit costs in a GED-consistent manner, predicting a zero cost for the aligned parts and the true GED for the unaligned parts, thereby eliminating spurious signals.
  
Our contributions are summarized as follows:
\begin{itemize}
    \item We identify a mismatch between existing GNN-based GSC methods and the core principle of GED, highlighting limitations in global alignment and cost attribution.
    \item We propose \textbf{GCGSim}, a novel end-to-end framework that is consistent with the GED mechanism by learning from the perspective of graph-level matching and substructure-level edit costs.
    \item We design three core technical modules: \textbf{GNCM} to 
    learn pair-aware graph representations, \textbf{PSGD} to perform principled disentanglement of substructures, and \textbf{IIR} to enforce semantic consistency for aligned embeddings 
    \item We conduct extensive experiments on four real-world datasets, demonstrating that GCGSim achieves state-of-the-art performance and validating the effectiveness of principled disentangled representation learning. 
\end{itemize}

\section{Preliminary}

\subsection{Graph Similarity Computation}
Given a graph $G=\left( V,E,X,A \right)$, where $V=\{ v^{\left( k \right)} \} _{k=1}^{N}$ is the set of nodes with number $N$, $E\subseteq  V\times V$ is the set of edges, $X\in \mathbb{R}^{N\times d}$ is the node features (e.g., labels or attribute vectors) with dimension $d$, and $A\in \mathbb{R} ^{N\times N} $ is the adjacency matrix, where $A\left[ k,l \right]=1$ if and only if edge $e^{\left( k,l \right) } \in E$, otherwise $A\left[ k,l \right] =0$. Given a graph database $\mathcal{D}$ and a set of query graphs $\mathcal{Q}$, the aim of GSC is to produce a similarity score $\mathcal{S} \in \mathbb{R} ^{\left( 0,1 \right]}$ between $\forall G_i\in \mathcal{D}$ and $\forall G_j\in \mathcal{Q}$, i.e., $\hat{s}_{ij}=S\left( G_i,G_j \right)$. $\hat{s}_{ij}$ is the predicted graph similarity score, $S:\mathcal{D} \times \mathcal{Q} \rightarrow \mathcal{S}$ is a trainable estimation model. We train the model based on a set of training triplet $\left( G_i,G_j,s_{ij} \right) \in \mathcal{D} \times \mathcal{Q} \times \mathcal{S}$ consisting of the input graph pairs and the ground-truth similarities $s_{ij}$. $s_{ij}$ is computed as follows:
\begin{equation}
    s_{ij}=\exp \left( -\frac{\text{GED}\left( G_i, G_j \right)}{\left(  N_i + N_j \right) /2} \right)
\end{equation}
where $\text{GED}\left( G_i, G_j \right)$ is ground-truth GED of  $G_i$ and $G_j$.

\subsection{Graph Edit Distance}\label{sec:GED}
\begin{definition}[Graph Edit Distance]
Given a graph pair $\left( G_i, G_j\right)$, the minimum cost of edit operations that transform  $G_i$ to $G_j$ is called the \textbf{graph edit distance} and denoted by $GED\left( G_i, G_j \right)$. Specifically, there are three types of edit operations: adding or removing an edge; adding or removing a node; and relabeling a node.
\end{definition}
The computation of GED is formulated as a minimization problem over all possible mappings $\pi$:
\begin{equation}
\begin{split}
     \text{GED}&\left( G_i, G_j \right) =\min\limits_{\pi}\text{ged}\left( G_i, G_j, \pi \right) \\
 & = \min\limits_{\pi} \text{ged}_V\left( G_i, G_j, \pi \right) + \text{ged}_E\left( G_i, G_j, \pi \right)
\end{split}
\end{equation}
where $\text{ged}\left( G_i, G_j, \pi \right)$ denotes the cost of edit operations that transform $G_i$ to $G_j$ under $\pi$, $\pi$ is an orthogonal permutation matrix that delineates the injective mapping between entities (nodes, edges) of $G_i$ and $G_j$. $\text{ged}_V$ and $\text{ged}_E$ respectively denote the cost of editing operations acting on nodes and edges, calculated as follows:
\begin{equation}
    \text{ged}_V\left( G_i, G_j, \pi \right) =\sum_{k=1}^{N}{\mathbb{I} \big( \left\| \left( X_i-\pi X_j \right) \left[ k \right] \right\| _p \ne 0 \big)}
\end{equation}
\begin{equation}
    \begin{split}
    \text{ged}_E&\left( G_i, G_j, \pi \right) \\ &=0.5\times \sum_{k=1}^{N}{\sum_{l=1}^{N}{\mathbb{I} \left( \left( A_i-\pi A_j\pi^T \right) \left[ k,l \right] \ne 0 \right)}}
    \end{split}
\end{equation}
where $\mathbb{I}\left( \cdot \right)$ is the indicator function that returns 1 if the condition holds and 0 otherwise, $\left\| \cdot \right\| _p$ denotes $p$-norm, $\pi X_j$ is the row-permuted transformation of $X_j$, $\pi A_j \pi^T$ is the row- and column-permuted transformation of $A_j$. Note that if $G_i$ and $G_j$ have different numbers of nodes, pad the smaller $X$ and $A$ with zeros to make the two graphs equal in size.
\begin{definition}[Optimal Alignment]
Given a graph pair $\left( G_i, G_j\right)$, $\pi^*$ denotes the \textbf{optimal alignment} between $G_i$ and $G_j$, if
\begin{equation}
    \pi^* = arg \min \limits_{\pi} \text{ged}\left( G_i, G_j, \pi \right)
\end{equation}
\end{definition}
\begin{definition}[Aligned and Unaligned substructure] Given a graph pair $\left( G_i, G_j\right)$, the \textbf{aligned substructure} of $G_i$ with respect to $G_j$ is defined as a 2-tuple $as_i = \left( V_i^{as}, E_i^{as}\right)$, where
\begin{equation}
V_i^{as}=\big\{ v^{\left( k \right)}_i \,|\,\left\| \left( X_i-\pi^*X_j \right) \left[ k \right] \right\| _p = 0 ,\, v^{\left( k \right)}_i\in V_i \big\} 
\end{equation}
and
\begin{equation}
E_i^{as}=\big\{ e^{\left( k,l \right)}_i \,|\, \big( A_i-\pi^*A_j{\pi^*}^T \big) \left[ k,l \right] = 0 ,\, e^{\left( k,l \right)}_i\in E_i \big\}. 
\end{equation}
the \textbf{unaligned substructure} of $G_i$ with respect to $G_j$ is defined as $us_i = \left( V_i^{us}, E_i^{us}\right)$, where
\begin{equation}
V_{i}^{us}=\big\{  v^{\left( k \right)}_i\,|\, v^{\left( k \right)}_i\in V_i \,\, and \,\,  v^{\left( k \right)}_i\notin V_{i}^{as} \big\}
\end{equation}
and
\begin{equation}
E_{i}^{us}=\big\{ e^{\left( k,l \right)}_i \,|\, e^{\left( k,l \right)}_i\in E_i  \,\, and \,\, e^{\left( k,l \right)}_i\notin E_{i}^{as} \big\}.
\end{equation}
\end{definition}

For a graph pair $\left( G_i, G_j \right)$, where $G_i=\left( V_i, E_i\right)$ and $G_j=\left( V_j, E_j\right)$, Fig.~\ref{fig:GED} examples a graph edit path that transform $G_i$ to $G_j$ with minimum cost and visualizes the partition of aligned substructures and unaligned substructures. Specifically, the aligned substructures of $G_i$ and $G_j$ are defined as
$as_i=\big( \big\{ v_{i}^{\left( 1 \right)},v_{i}^{\left( 2 \right)},v_{i}^{\left( 3 \right)} \big\} ,\big\{ e_{i}^{\left( 1,2 \right)},e_{i}^{\left( 2,3 \right)},e_{i}^{\left( 3,4 \right)} \big\} \big) $ and
$as_j=\big( \big\{ v_{j}^{\left( 1 \right)},v_{j}^{\left( 4 \right)},v_{j}^{\left( 5 \right)} \big\} ,\big\{ e_{j}^{\left( 1,5 \right)},e_{j}^{\left( 4,5 \right)},e_{j}^{\left( 3,4 \right)} \big\} \big) $, respectively. The unaligned substructures are
$us_i=\big( \big\{ v_{i}^{\left( 4 \right)} \big\} ,\big\{ e_{i}^{\left( 1,4 \right)} \big\} \big) $, 
$us_j=\big( \big\{ v_{j}^{\left( 2 \right)},v_{j}^{\left( 3 \right)} \big\} ,\big\{ e_{j}^{\left( 2,3 \right)},e_{j}^{\left( 2,5 \right)} \big\} \big) $. The optimal alignment $\pi^*$ reflects a node mapping
$\big( v_{i}^{\left( 1 \right)}\mapsto v_{j}^{\left( 1 \right)},v_{i}^{\left( 2 \right)}\mapsto v_{j}^{\left( 5 \right)},v_{i}^{\left( 3 \right)}\mapsto v_{j}^{\left( 4 \right)} \big) $ and a edge mapping
$\big( e_{i}^{\left( 1,2 \right)}\mapsto e_{j}^{\left( 1,5 \right)},e_{i}^{\left( 2,3 \right)}\mapsto e_{j}^{\left( 4,5 \right)},e_{i}^{\left( 3,4 \right)}\mapsto e_{j}^{\left( 3,4 \right)} \big) $. 

\section{Related work}
Research on the GED approximation can be divided into two main streams: 
heuristic algorithms and 
learning-based methods 

\subsection{Heuristic Methods}
Traditional approaches for approximating GED in polynomial time often rely on combinatorial heuristics. For instance, A*-Beam~\cite{b20} prunes the search space of the exact A*  algorithm \cite{b19} with a user-defined beam width to balance accuracy and efficiency. A more prevalent family of methods models GED computation as a Linear Sum Assignment Problem (LSAP). 
The Hungarian method~\cite{b21}, a cornerstone of this approach, first constructs a cost matrix where each entry estimates the cost of matching two nodes. The problem is then solved using the Hungarian algorithm~\cite{kuhn1955hungarian} to find a minimum-cost node assignment, which serves as the GED approximation. Subsequent works, such as VJ~\cite{fankhauser2011speeding}, have focused on improving the computational efficiency of solving this assignment problem. While effective, the performance of these heuristic methods is fundamentally limited by their reliance on hand-crafted rules for defining the cost matrix, which may struggle to capture complex structural patterns.

\subsection{GNN-based Methods}
To overcome the limitations of hand-crafted heuristics, recent work has shifted towards learning-based approaches, which can be broadly categorized into two main paradigms.

\paragraph{Graph-level Embedding Regression}
This paradigm first maps each graph 
into a single, fixed-size vector representation. The final similarity score is then computed directly from these two graph embeddings, typically using a simple regressor.
Representative works 
include SimGNN~\cite{b9}, EGSC~\cite{b24}, N2AGim~\cite{b25}, and GraSP~\cite{zheng2025grasp}.
While computationally efficient, this paradigm tends to lose fine-grained structural details, as the entire graph's information is compressed into a single vector, thereby limiting its expressive power for the correspondence-rich task of GED calculation.

\paragraph{Node-level Interaction Matching}
To capture more detailed structural alignments, another powerful paradigm is based on node-level interaction.
These methods also begin by learning node embeddings for each graph but then introduce an interaction module. The core idea is to learn a pairwise interaction matrix that explicitly encodes the alignment information or edit costs between nodes
, which is then aggregated to predict the final score. 
Prominent examples include GraphSim~\cite{b10}, NA-GSL~\cite{b25}, GraphOTSim~\cite{b43}, GEDGNN~\cite{b42} and GEDIOT~\cite{cheng2025computing}. These approaches employ diverse mechanisms, such as applying CNNs to the similarity matrix or using optimal transport and attention to learn soft correspondences.

Our work is situated within this 
node-level interaction paradigm. Nevertheless, we argue that even these methods often exhibit a subtle mismatch with the combinatorial nature of GED. First, they tend to rely on local neighborhood information when computing node similarities, struggling to capture the optimal alignment that GED seeks. Second, by learning from a dense similarity (or cost) matrix, they conflate the true edit costs derived from \textit{unaligned} substructures with irrelevant signals from 
aligned parts. Our proposed framework addresses these specific limitations by reformulating the interaction process to be more consistent with the principles of 
global 
matching and edit cost attribution inherent to GED.

\begin{figure*}[t]
  \centering
  \includegraphics[width=0.98\linewidth]{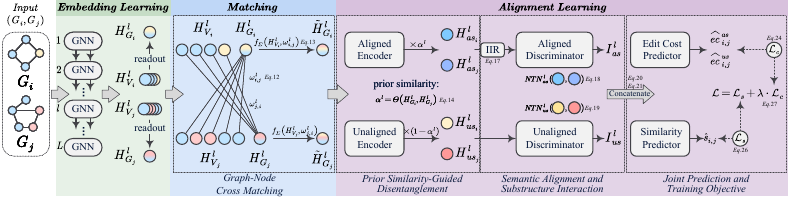} 
  \caption{The architecture of GCGSim}
  \label{overview}
\end{figure*}

\section{Method}
The architecture of GCGSim is illustrated in Fig.~\ref{overview}.
The entire process is composed of three 
stages: (1) \textbf{Embedding Learning}, where we generate 
multi-scale representations for each graph in the pair; (2) \textbf{Matching}, where a 
graph-level matching module creates pair-aware contextualized graph embeddings; and (3) \textbf{Alignment Learning}, the core of our method, which first disentangles the embeddings into representations for aligned and unaligned substructures and then jointly predicts the final similarity and edit costs. We will elaborate on each of these stages in the subsequent sections.

\subsection{Embedding Learning}
\label{sec:embedding_learning}
The initial stage of our framework is dedicated to generating rich, multi-scale representations for each graph in an input pair $(G_i, G_j)$. By leveraging the hierarchical outputs of a deep GNN, we capture structural information across varying receptive fields. We employ a Siamese architecture, where an identical encoder with shared parameters processes each graph.

\subsubsection{Node Embedding Learning}
Our node encoder is built upon an $L$-layer GNN to iteratively refine node features. As the GNN layers deepen, each node's receptive field expands, allowing the model to capture progressively larger-scale structural patterns. We select the Residual Gated Graph Convolutional Network (RGGC)~\cite{bresson2017residual} as our backbone for its enhanced expressive power, which uses gating mechanisms and residual connections to effectively capture complex patterns. The update rule for a node $k$'s representation at the $l$-th layer is:
\begin{equation}
\label{eq:rggc_update}
\begin{split}
    H_V^{l}&[k] = H_V^{l-1}[k] \\ &+ \text{ReLU} \big( W_S H_V^{l-1}[k] + \sum_{u \in \mathcal{N}(k)} \eta_{k,u} \odot W_N H_V^{l-1}[u] \big)
\end{split}
\end{equation}
where $H_V^{l}$ denotes the matrix of node embeddings at the $l$-th layer, $W_S$ and $W_N$ are learnable matrices, $\mathcal{N}(k)$ is the neighborhood of node $k$, $\odot$ is the Hadamard product, and $\eta_{k,u}$ is the learned gate that modulates the message from node $u$ to $k$. This RGGC encoder, with its weights shared, is applied independently to both $G_i$ and $G_j$. The process yields a full sequence of layer-wise node representations $\{H_{V_i}^{0}, \dots, H_{V_i}^{L}\}$ for $G_i$ and $\{H_{V_j}^{0}, \dots, H_{V_j}^{L}\}$ for $G_j$, where each matrix $H_V^{l}$ captures structural patterns at a distinct scale.

\subsubsection{Graph Embedding Learning}
Following the generation of multi-scale node embeddings, we derive a corresponding set of graph-level representations. It is crucial that features from different scales are aggregated consistently. Therefore, at each layer $l$, we apply a shared, permutation-invariant readout function to the node embeddings $H_{V_i}^{l}$ and $H_{V_j}^{l}$ respectively. We employ DeepSets~\cite{zaheer2017deep} for this purpose:
\begin{equation}
    H_{G}^{l} = \text{MLP}_{\text{DS}}^{l} \big( \sum_{k=1}^{N} H_{V}^{l}[k] \big)
    \label{eq:graph_embedding_layerwise}
\end{equation}
The DeepSets module, specifically its $\text{MLP}_{\text{DS}}^{l}$, shares parameters for both graphs, ensuring a uniform mapping from node sets to graph embeddings. This layer-wise aggregation results in a sequence of graph embeddings $\{H_{G_i}^{0}, \dots, H_{G_i}^{L}\}$ for $G_i$ and a corresponding sequence $\{H_{G_j}^{0}, \dots, H_{G_j}^{L}\}$ for $G_j$. 

\subsection{Matching}
\subsubsection{Graph-Node Cross Matching}
The node and graph embeddings, $H_{V_i}^l$ and $H_{G_i}^l$, derived in the previous stage, encapsulate multi-scale structural information. However, they are 
context-agnostic, as the representation of each graph is learned in isolation, without any information from the other graph in the pair. This intrinsic limitation hinders the model from directly reasoning about the alignment between the two graphs.

To address this, we introduce the Graph-Node Cross Matching (GNCM), a mechanism designed to generate \textit{pair-aware} contextualized graph representations. The core rationale is that while a node embedding $H_{V_i}^{l}[k]$ captures the intrinsic properties of its local substructure, its contribution to the final similarity score is not absolute. Instead, its significance must be modulated by the global context of the graph it is being compared to $G_j$. For instance, a complex substructure in $G_i$ may contribute little to the similarity if no analogous structure exists within $G_j$.

Conceptually, GNCM functions as a cross-graph attention mechanism. It learns to dynamically re-weight the importance of each local substructure in one graph based on its relevance to the global 
context of the other. This allows the model to emphasize structurally congruent substructures and de-emphasize dissimilar ones, thereby calibrating the representations for the specific comparison task.

Formally, we first compute the node-wise relevance scores $\omega_{i,j}^l$ by measuring the correspondence between each node embedding in $G_i$ and the global graph embedding of $G_j$:
\begin{equation}
    \omega_{i,j}^{l}[k] = \Theta \left( H_{V_i}^{l}[k], H_{G_j}^{l} \right)
    \label{eq:weight}
\end{equation}
where $\omega_{i,j}^{l}[k]$ is the attention score for the $k$-th node, and $\Theta(\cdot, \cdot)$ is the cosine similarity function. 
Subsequently, these scores are used to produce a pair-aware graph embedding $\tilde{H}_{G_i}^{l}$ via a weighted aggregation of the node embeddings:
\begin{equation}
    \tilde{H}_{G_i}^{l} = f_{\varSigma}\left( \omega _{i,j}^l, H_{V_i}^l \right)= \sum_{k=1}^{N} \omega_{i,j}^{l}[k] \cdot H_{V_i}^{l}[k]
    \label{eq:attentive}
\end{equation}
The resulting embedding, $\tilde{H}_{G_i}^{l}$, represents a contextualized view of graph $G_i$, re-weighted according to the structure of $G_j$. A symmetric operation is performed to compute $\tilde{H}_{G_j}^{l}$. 

\subsection{Alignment Learning}
\subsubsection{Prior Similarity-Guided Disentanglement}
\label{sec:psgd}

After obtaining the pair-aware graph embeddings for both graphs in the pair, $\tilde{H}_{G_i}^l$ and $\tilde{H}_{G_j}^l$, which encapsulate holistic information about their relationship, our next crucial step is to disentangle these representations. To this end, each embedding is separately processed to yield representations for its respective \textit{aligned} and \textit{unaligned} substructures, using two distinct modules: an Aligned Encoder and an Unaligned Encoder.

A core challenge lies in determining the relative contribution of each substructure type to the overall graph representation. Intuitively, if two graphs are highly similar, their aligned substructures should dominate the final similarity score. Conversely, for dissimilar graphs, the unaligned substructures and their associated edit costs are more informative. To operationalize this intuition, we propose using the similarity of the graph embeddings themselves, which serves as a task-aware prior, to dynamically estimate the influence weights for disentanglement. This prior, denoted as $\alpha$, is computed as:
\begin{equation}
    \alpha^l = \Theta (H_{G_i}^{l}, H_{G_j}^{l})
    \label{eq:alpha_prime}
\end{equation}
where $\Theta(\cdot, \cdot)$ is the cosine similarity function.

This choice is not merely a heuristic. In Section~\ref{sec:theorypsgd}, we provide a rigorous theoretical justification from a variational inference perspective, demonstrating that using graph embedding similarity as the prior is a principled approach to optimize the evidence lower bound (ELBO).

With this principled weight $\alpha^{l}$, we can now guide the disentanglement process. The Aligned Encoder and Unaligned Encoder, both parameterized as MLPs, generate the aligned embedding $H_{as_i}^l$ and unaligned embedding $H_{us_i}^l$ at layer $l$ as follows:
\begin{align}
    H_{as_i}^l = \alpha^{l} \text{MLP}_{as}^{{l}}(\tilde{H}_{G_i}^l) \label{eq:has}
\end{align}
\begin{align}
    H_{us_i}^l = (1 - \alpha^{l})
    \text{MLP}_{us}^{{l}}(\tilde{H}_{G_i}^l) \label{eq:hus}
\end{align}
where $\text{MLP}_{as}^{l}$ and $\text{MLP}_{us}^{l}$ are the respective encoders. The embeddings for the graph $G_j$ ($H_{as_j}^l$ and $H_{us_j}^l$) are obtained through a symmetric operation. 

\subsubsection{Semantic Alignment and Substructure Interaction}
\label{sec:interaction_fusion}
Having disentangled the representations, the next step is to interact them to learn their correspondence and fuse the multi-layer information.
\paragraph{Semantic Alignment via Intra-Instance Replicate (IIR)}
A prerequisite for a meaningful interaction between $H_{as_i}^{l}$ and $H_{as_j}^{l}$ is that they must share the same canonical semantics of "aligned-ness". Directly enforcing this with typical regularization constraints can interfere with the primary GSC task \cite{b37, sanchez2020learning}. We therefore introduce a novel consistency-based method, Intra-Instance Replicate (IIR). 

The intuition is to treat this as a data augmentation task in the embedding space. During training, for a given pair $(G_i, G_j)$, we randomly replace the original $H_{as_i}^{l}$ with a convex combination of itself and its counterpart $H_{as_j}^{l}$:
\begin{equation}
    \hat{H}_{as_i}^{l} = \tau H_{as_i}^{l} + (1-\tau) H_{as_j}^{l}
    \label{eq:IIR}
\end{equation}
where $\tau \sim \text{Bernoulli}(\beta)$ with $\beta \in [0,1]$ being a hyper-parameter. The model is then tasked with making a prediction using this augmented embedding $\hat{H}_{as_i}^{l}$. Since the ground-truth label remains unchanged regardless of whether this augmentation occurs, the model is implicitly forced to be robust to this replication. 

To minimize the long-term training objective, the optimal strategy for the encoders is to learn representations $H_{as_i}^{l}$ and $H_{as_j}^{l}$ that are semantically interchangeable. By making the representations nearly identical, the model effectively learns that the useful signal is the structural pattern that is common to both. 

\paragraph{Substructure-level Interaction}
With semantically aligned representations, we now perform interaction using the Neural Tensor Network (NTN) \cite{b9}, chosen for its efficacy in modeling complex relationships between embeddings \cite{b23}. 
Two distinct NTN modules function as the Aligned Discriminator and Unaligned Discriminator.
For each layer $l$, we compute interaction embeddings for both aligned and unaligned substructures:
\begin{align}
    I_{as}^l &= \text{NTN}_{as}^{l}( \hat{H}_{as_i}^{l}, H_{as_j}^{l}) \label{eq:ntnas} \\
    I_{us}^l &= \text{NTN}_{us}^{l}( H_{us_i}^{l}, H_{us_j}^{l}) \label{eq:ntnus}
\end{align}
where $\hat{H}_{as_i}^{l}$ is the potentially augmented aligned embedding from \eqref{eq:IIR} (if no augmentation is applied, $\hat{H}_{as_i}^{l}=H_{as_i}^{l}$).

To leverage the structural information captured across all $L$ layers, we concatenate the layer-wise interaction embeddings to form final, comprehensive representations for the aligned and unaligned substructures:
\begin{align}
    I_{as} = I_{as}^1 \parallel I_{as}^2 \parallel \dots \parallel I_{as}^L \label{eq:fusion_as} \\
    I_{us} = I_{us}^1 \parallel I_{us}^2 \parallel \dots \parallel I_{us}^L \label{eq:fusion_us}
\end{align}
where $\parallel$ denotes the concatenation operator. These fused representations, $I_{as}$ and $I_{us}$, now serve as the input for the final predictors.

\subsubsection{Joint Prediction and Training Objective}
\label{sec:prediction_loss}
The final stage of our framework involves two predictors that jointly learn from the fused interaction embeddings.

\paragraph{Edit Cost Prediction (ECP)}
The Edit Cost Predictor computes the edit costs directly from the fused substructure representations. The aligned substructures should, by definition, have an edit cost of zero, while the unaligned substructures are responsible for the total GED. This is formalized as:
\begin{equation}
    \widehat{ec}_{ij}^{as} = \text{MLP}_{as}^{ec}(I_{as})
\end{equation}
\begin{equation}
    \widehat{ec}_{ij}^{us} = \text{MLP}_{us}^{ec}(I_{us})
\end{equation}
The corresponding Mean Squared Error (MSE) loss function is:
\begin{equation}
    \mathcal{L}_c=\frac{1}{n}\sum_{(G_i,G_j)}{\left[ (0-\widehat{ec}_{ij}^{as})^2 + (\text{GED}(G_i,G_j) - \widehat{ec}_{ij}^{us})^2 \right]}
    \label{eq:loss_c}
\end{equation}

\paragraph{Similarity Prediction}
The Similarity Predictor computes the final graph similarity score. It takes the concatenated aligned and unaligned interaction embeddings as input, allowing it to consider information from both sources:
\begin{equation}
    \hat{s}_{ij} = \text{MLP}_{\text{sim}}(I_{as} \parallel I_{us})
\end{equation}
This is trained using a standard MSE loss against the ground-truth similarity $s_{ij}$:
\begin{equation}
    \mathcal{L}_s =\frac{1}{n}\sum_{(G_i,G_j)}{(s_{ij}-\hat{s}_{ij})^2}
    \label{eq:loss_s}
\end{equation} 

\paragraph{Final Objective}
The model is trained end-to-end by minimizing a weighted sum of the two loss components:
\begin{equation}
    \mathcal{L} = \mathcal{L}_{s} + \lambda \cdot \mathcal{L}_{c}
    \label{eq:final_loss}
\end{equation}
where $\lambda$ is a hyper-parameter that balances the contribution of the edit cost prediction task.

\begin{table*}[bhtp]
\caption{Dataset statistics.}
\label{tab_data}
\begin{center}
    \resizebox{0.85\linewidth}{!}{
    \begin{tabular}{ccccccc}
    \toprule
    \textbf{Dataset} & \textbf{Graph Meaning} & \textbf{\#Graph} & \textbf{\#Pairs} & \textbf{AVG \#Nodes} & \textbf{AVG \#Edges} & \textbf{\#Features} \\
    \midrule
    LINUX & Program Dependency Graphs & 1000 & 1M & 7.6 & 6.9 & 1 \\
    AIDS700nef & Anti-HIV  Compounds & 700 & 490K & 8.9 & 8.8 & 29 \\
    IMDBMulti & Actor/Actress Ego-Networks & 1500 & 2.25M & 13.0 & 65.9 & 1 \\
    PTC & Carcinogenic Compounds & 344 & 1.90K & 25.5 & 51.9 & 19 \\
    \bottomrule
    \end{tabular}}
\end{center}
\end{table*}

\section{Theoretical Analysis of Prior-Guided Disentanglement}\label{sec:theorypsgd}

In this section, we provide a formal theoretical justification for our Prior Similarity-Guided Disentanglement (PSGD) mechanism. We frame the task from the perspective of variational inference on a probabilistic latent variable model. 
This rigorous approach demonstrates that our design---which employs the similarity of graph embeddings as a dynamic, data-dependent prior---is a principled strategy to optimize the evidence lower bound (ELBO), rather than a mere heuristic.

\subsection{Probabilistic Modeling Framework}

We 
formulate the graph similarity prediction task as a probabilistic generative model. Given a pair of graphs $(G_i, G_j)$, our objective is to learn a model, parameterized by $\theta$, that captures the conditional probability $p_\theta(s_{ij} | G_i, G_j)$ of the true similarity score $s_{ij}$. We introduce a discrete latent variable $k \in \{as, us\}$, which represents the underlying structural source contributing to the final similarity, where $as$ denotes the \textbf{aligned substructure} and $us$ denotes the \textbf{unaligned substructure}.

Under our generative assumption, the marginal likelihood of the similarity score $s_{ij}$ is obtained by marginalizing out the latent variable $k$:
\begin{equation}
p_\theta(s_{ij} | G_i, G_j) = \sum_{k \in \{as, us\}} p_\theta(s_{ij} | G_i, G_j, k) \, p_\theta(k | G_i, G_j)
\label{eq:marginal_likelihood}
\end{equation}
Here, $p_\theta(k | G_i, G_j)$ is the \textbf{prior distribution} over the substructure types, modeled as a Bernoulli distribution over the set $\{as, us\}$. The conditional likelihood $p_\theta(s_{ij} | G_i, G_j, k)$ dictates the contribution of each substructure type. The learning objective is to maximize the log-marginal likelihood, $\log p_\theta(s_{ij} | G_i, G_j)$.

\subsection{Variational Formulation and Justification}
Directly optimizing the objective in ~\eqref{eq:marginal_likelihood} is intractable. 
We therefore employ variational inference to maximize its ELBO.

\begin{lemma}[The Evidence Lower Bound Objective]
\label{lemma:elbo}
The log-marginal likelihood admits the following ELBO,
\begin{align}
\log p_\theta(s_{ij} | G_i, G_j) \ge \mathcal{L}(\theta, \phi) ={} & \mathbb{E}_{q_\phi(k)} [\log p_\theta(s_{ij} | G_i, G_j, k)] \nonumber \\
& - D_{KL}(q_\phi(k) \,||\, p_\theta(k)) \label{eq:elbo}
\end{align}
where for brevity, $q_\phi(k) \triangleq q_\phi(k | G_i, G_j, s_{ij})$ and $p_\theta(k) \triangleq p_\theta(k | G_i, G_j)$, $q_\phi(k | G_i, G_j, s_{ij})$ is a variational distribution, parameterized by $\phi$, that approximates the true posterior $p_\theta(k | G_i, G_j, s_{ij})$.
\end{lemma}
\begin{proof}[Proof]
\begin{equation}
\resizebox{\linewidth}{!}{$
\begin{aligned}
     &\log p_{\theta}\bigl( s_{ij}|G_i,G_j \bigr) \\
     &=\log \sum_k{p_{\theta}\bigl( s_{ij},k|G_i,G_j \bigr)}
    \\
    &=\log \sum_k{q_{\phi}\bigl( k|G_i,G_j,s_{ij} \bigr) \frac{p_{\theta}\bigl( s_{ij},k|G_i,G_j \bigr)}{q_{\phi}\bigl( k|G_i,G_j,s_{ij} \bigr)}}
    \\
    & \geqslant \sum_k{q_{\phi}\bigl( k|G_i,G_j,s_{ij} \bigr) \log \frac{p_{\theta}\bigl( s_{ij},k|G_i,G_j \bigr)}{q_{\phi}\bigl( k|G_i,G_j,s_{ij} \bigr)}} 
    \\
    & \quad {} \left( \text{by}\,\,\text{Jensen's}\,\,\text{Inequality} \right)  
    \\
    &=\sum_k{q_{\phi}\bigl( k|G_i,G_j,s_{ij} \bigr) \log \frac{p_{\theta}\bigl( s_{ij}|G_i,G_j,k \bigr) p_{\theta}\bigl( k|G_i,G_j \bigr)}{q_{\phi}\bigl( k|G_i,G_j,s_{ij} \bigr)}}
    \\
    &=\sum_k{q_{\phi}\bigl( k|G_i,G_j,s_{ij} \bigr) \log p_{\theta}\bigl( s_{ij}|G_i,G_j,k \bigr)} 
    \\
    & \quad {} -\sum_k{q_{\phi}\bigl( k|G_i,G_j,s_{ij} \bigr) \log \frac{q_{\phi}\bigl( k|G_i,G_j,s_{ij} \bigr)}{p_{\theta}\bigl( k|G_i,G_j \bigr)}}
    \\
    &=\mathbb{E} _{q_{\phi}\left( k|G_i,G_j,s_{ij} \right)}\left[ {\log p_{\theta}\left( s_{ij}|G_i,G_j,k \right)} \right]  
    \\
    & \quad {} -D_{KL}\left( q_{\phi}\bigl( k|G_i,G_j,s_{ij} \bigr) \parallel p_{\theta}\bigl( k|G_i,G_j \bigr) \right) 
    \label{eq18}
\end{aligned}
$}
\end{equation}
\end{proof}
The optimization of the ELBO (Equation.~\eqref{eq:elbo}) comprises two objectives:
1) Maximizing the \textit{reconstruction term}, which is directly handled by our model's MSE loss functions.
2) Minimizing the \textit{KL-divergence}.
Our theoretical argument centers on a principled design of the prior $p_{\theta}\left( k|G_i,G_j \right)$ to effectively minimize this term.


\begin{lemma}[Monotonicity of the Ideal Posterior]
\label{lemma:posterior}
The ideal posterior probability $p_\theta(k=as | G_i, G_j, s_{ij})$, which represents the probability of the substructure being aligned given knowledge of the true similarity $s_{ij}$, is a monotonically increasing function of $s_{ij}$.
\end{lemma}
\begin{proof}
This lemma follows from the definition of graph similarity. As $s_{ij} \to 1$, the graphs approach isomorphism, implying that the vast majority of their structure is aligned; hence, $p_{\theta}(k=as|\cdot)$ must approach 1. Conversely, as $s_{ij} \to 0$, the graphs are structurally disparate, dominated by unaligned components, causing $p_{{\theta}}(k=as|\cdot)$ to approach 0. The posterior probability is therefore monotonically correlated with $s_{ij}$.
\end{proof}

\begin{theorem}[Optimizing the ELBO via an Informed Prior Design]
\label{theorem:main}
The KL-divergence term in the ELBO is effectively minimized by designing a prior distribution $p_\theta(k | G_i, G_j)$ that aligns with the target behavior of the variational posterior $q_\phi(k)$. The similarity of graph embeddings, $\Theta(H_{G_i}, H_{G_j})$, serves as a principled function for such a prior.
\end{theorem}
\begin{proof}
The core of variational inference is to optimize the variational distribution $q_\phi(k)$ to be the closest possible approximation of the ideal posterior $p_\theta(k|G_i,G_j,s_{ij})$. Consequently, the variational objective compels an optimized $q_\phi(k)$ to learn and exhibit the properties of this ideal posterior. From Lemma~\ref{lemma:posterior}, we know a key property of the ideal posterior is its monotonic dependence on the true similarity score $s_{ij}$. Therefore, the learning target for $q_\phi(k)$ is to acquire this same monotonic behavior.

Our strategy is to design a prior $p_\theta(k)$ that preemptively matches this learning target of $q_\phi(k)$. This requires a function that mimics the behavior of $s_{ij}$ but adheres to a critical constraint: the prior must not be a direct function of the label $s_{ij}$ to prevent information leakage.

This challenge motivates the use of a proxy. Within our GNN framework, the graph embeddings $H_{G_i}$ and $H_{G_j}$ are representations learned specifically to capture the structural information relevant for similarity prediction. Thus, their similarity, $\Theta(H_{G_i}, H_{G_j})$, constitutes the model's own best, label-independent estimate of $s_{ij}$ and, by extension, the behavior of the ideal posterior.

Based on this, we set the prior probability for the aligned substructure as:
\begin{equation}
\alpha := p_\theta(k=as | G_i, G_j) = \Theta(H_{G_i}, H_{G_j})
\label{eq:prior_as}
\end{equation}
Correspondingly, as the latent space for $k$ is binary and its distribution must sum to unity, the prior probability for the unaligned substructure is complementarily defined as:
\begin{equation}
p_\theta(k=us | G_i, G_j) = 1 - \alpha = 1 - \Theta(H_{G_i}, H_{G_j})
\label{eq:prior_us}
\end{equation}

This formulation is not an ad-hoc choice. It establishes a data-dependent prior, a standard technique in Amortized Variational Inference. 
This informed prior is therefore constructed to match the learning target of the variational posterior $q_\phi(k)$.
Such a design preemptively satisfies the objective of minimizing the KL-divergence, which permits the model to dedicate its capacity to the reconstruction task.
\end{proof}


\begin{table*}[tbp]
\caption{Main result of GCGSim and baseline models on benchmark datasets. Bold means the best result and underline means the second result.}
\label{tab_1}
\begin{center}
    \resizebox{0.865\linewidth}{!}{
    \begin{tabular}{ccccccccccc}
    \toprule
    \multirow{2}{*}{\textbf{method}} & \multicolumn{5}{c}{\textbf{AIDS700nef}} & \multicolumn{5}{c}{\textbf{LINUX}}\\
    \cmidrule(lr){2-6} \cmidrule(lr){7-11} 
    & MSE$\downarrow $ & $\rho \uparrow $ & $\tau \uparrow$ & $p@10 \uparrow $ & $p@20 \uparrow $ & MSE$\downarrow $ & $\rho \uparrow$ & $\tau \uparrow$ & $p@10 \uparrow$ & $p@20 \uparrow$ \\
    \midrule
    SimGNN  & 1.573 & 0.835 & 0.678 & 0.417 & 0.489 & 2.479 & 0.912 & 0.791 & 0.635 & 0.650  \\
    GraphSim & 2.014 & 0.839 & 0.662 & 0.401 & 0.499 & 0.762 & 0.953 & 0.882 & 0.956 & 0.951  \\
    GMN & 4.610 & 0.672 & 0.497 & 0.200 & 0.263 & 2.571 & 0.906 & 0.763 & 0.888 & 0.856 \\
    MGMN & 2.297 & 0.904 & 0.736 & 0.456 & 0.534 & 2.040 & 0.965 & 0.858 & 0.956 & 0.920 \\
    GEDGNN& 3.505 & 0.876& 0.762& 0.711& 0.736 & 1.763 & 0.960 & 0.903 & 0.962 & 0.976\\
    NA-GSL & 2.223 & 0.880 & 0.733 & 0.488 & 0.586 & 0.196 & \textbf{0.990} & \textbf{0.960} & 0.991 & 0.984 \\
    ERIC & 1.383 & 0.906& 0.740& 0.679 & 0.746 & 0.113& \underline{0.988} & 0.908 & \textbf{0.994} & \underline{0.996} \\
    GEDIOT & 1.348 & 0.902 & 0.733 & 0.744& 0.790 & 0.104 & 0.978 & \underline{0.928}& 0.983 & 0.990 \\ 
    GraSP & \underline{1.240} & \underline{0.912}& \underline{0.744}& \underline{0.756}& \underline{0.803}& \underline{0.097}& 0.974 & 0.923 & 0.986 & 0.994\\
    GCGSim & \textbf{1.069} & \textbf{0.923} &\textbf{0.772} & \textbf{0.767} & \textbf{0.815} & \textbf{0.066} & \textbf{0.990} & 0.915& \underline{0.993} & \textbf{0.997} \\ 
    \bottomrule
    \end{tabular}}
\end{center}

\begin{center}
    \resizebox{0.865\linewidth}{!}{
    \begin{tabular}{ccccccccccc}
    \toprule
    \multirow{2}{*}{\textbf{method}} & \multicolumn{5}{c}{\textbf{IMDBMulti}} & \multicolumn{5}{c}{\textbf{PTC}}\\
    \cmidrule(lr){2-6} \cmidrule(lr){7-11} 
    & MSE$\downarrow $ & $\rho \uparrow $ & $\tau \uparrow$ & $p@10 \uparrow $ & $p@20 \uparrow $ & MSE$\downarrow $ & $\rho \uparrow$ & $\tau \uparrow$ & $p@10 \uparrow$ & $p@20 \uparrow$ \\
    \midrule
    SimGNN  & 1.437 & 0.871 & 0.752 & 0.710 & 0.769 & 2.250 & 0.904 & 0.809 & 0.465 & 0.628 \\
    GraphSim & 1.924 & 0.825 & 0.821 & 0.813 & 0.825 & 3.482 & 0.825 & 0.712 & 0.418 & 0.535 \\
    GMN & 4.320 & 0.665 & 0.601 & 0.588 & 0.593 & 2.630 & 0.841 & 0.801 & 0.520 & 0.652 \\
    MGMN & 0.496 & 0.881 & 0.803 & 0.874 & 0.861 & 1.856 & 0.881 & 0.802 & 0.541 & 0.640 \\
    GEDGNN & 2.560 & 0.837 & 0.753& 0.792 & 0.834 & 1.704 & 0.913 & 0.796 & 0.543 & 0.659\\
    NA-GSL & 1.054 & 0.874 & 0.790 & 0.820 & 0.842 &  \multicolumn{5}{c}{OOM} \\
    ERIC & \textbf{0.385} & 0.890 & 0.791 & \underline{0.882} & 0.891 & 1.878 & 0.922 & 0.779 & 0.562 & 0.680 \\
    GEDIOT & 0.860 & 0.930 & \underline{0.838} & 0.879 & \underline{0.895} & 1.697 & 0.928 & 0.796 & 0.596 & 0.675 \\
    GraSP & \underline{0.393} & \underline{0.932} & 0.828 & 0.860 & 0.878 & \underline{1.673} & \underline{0.932} & \textbf{0.842} & \underline{0.612} & \underline{0.707} \\
    GCGSim & 0.564 & \textbf{0.949} & \textbf{0.848} & \textbf{0.892} & \textbf{0.905} & \textbf{1.548} & \textbf{0.948} & \underline{0.806} & \textbf{0.617} & \textbf{0.722} \\ 
    \bottomrule
    \end{tabular}}

\end{center}
\end{table*}

\section{Experiment}
To demonstrate the effectiveness of GCGSim, we perform extensive experiments on four real-world benchmark datasets. 
We first compare our model against a wide range of state-of-the-art GNN-based methods for graph similarity computation. 
Then, through a series of ablation studies and qualitative analyses, we systematically investigate the impact of our core technical modules and validate the design principles of our framework.

\subsection{Dataset}\label{datatset}
We evaluate GCGSim on LINUX~\cite{b38}, AIDS700nef~\cite{b38}, IMDBMulti~\cite{b39}, and PTC~\cite{toivonen2003statistical}. The statistical information of the dataset is presented in Table~\ref{tab_data}. 
We adopt the same data splits as \cite{b9}: 60\verb|%|, 20\verb|%|, and 20\verb|%| of all graphs are used as the training, validation, and testing sets, respectively. 
The ground-truth GEDs of graph pairs in AIDS700nef and LINUX are computed by the A* algorithm \cite{b19}. For IMDBMulti and PTC, we use the minimum of the results of approximate algorithms, Beam \cite{b20}, Hungarian \cite{b21}, and VJ \cite{b40} as ground-truth GEDs, following the way in \cite{b9}.

\subsection{Model settings and training}
GCGSim uses PyTorchGeometric for evaluation. We employ the Adam optimizer with a learning rate of 0.001 and set the batch size to 64. We adopt 4 layers of RGGC, with the number of feature channels defined as 64, 64, 32, and 16. We set $\beta=0.05$ and $\lambda=0.05$. We run 50 epochs on each dataset, performing validation after each epoch. Finally, the parameters that result in the smallest validation loss are selected to evaluate the test data. All experiments were conducted on a Linux server equipped with an Intel(R) Core(TM) i9-10900X CPU @ 3.70GHz and a single NVIDIA GeForce RTX 3090.

\subsection{Evaluation metric}
To comprehensively evaluate our model on the GSC task, the following five metrics are adopted to evaluate results for fair comparisons: \textbf{Mean Squared Error} (MSE) (in the format of $10^{-3}$), which measures the average squared differences between the predicted and the ground-truth similarity scores. \textbf{Spearman’s Rank Correlation Coefficient} ($\rho$) and \textbf{Kendall’s Rank Correlation Coefficient} ($\tau $) evaluate the ranking correlations between the predicted and the true ranking results. \textbf{Precision at $\boldsymbol{k}$} ($p@k$) where $k$ = 10, 20, which is the intersection of the top $k$ results of the prediction and the ground-truth. The smaller the MSE, the better the performance of models; for $\rho $, $\tau $, and $p@k$, the larger the better.

\begin{figure*}[ht]
	\centering
    \includegraphics[width=0.87\linewidth]{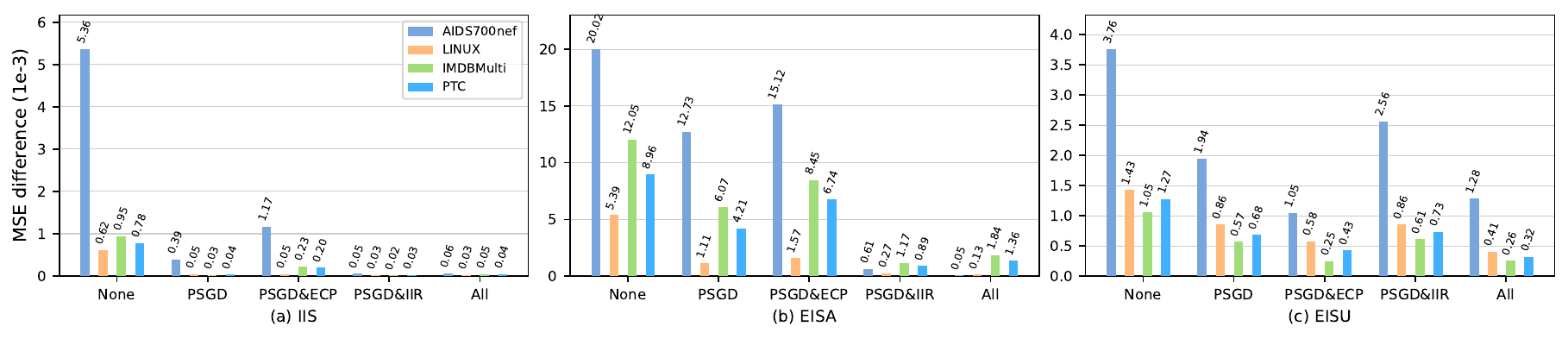}
\caption{
    Analysis of representation swapping. 
    The bars show the performance degradation (MSE difference) under three swapping settings: 
    (a) Intra-Instance Swap (IIS), 
    (b) Extra-Instance Swap for Aligned substructures (EISA), and 
    (c) Extra-Instance Swap for Unaligned substructures (EISU). 
    The x-axis represents different model variants, from a baseline ('None') to our full model ('All'), to show the contribution of each component to robustness. 
    A smaller bar indicates greater robustness.
}
    \label{fig:swap_experiments}
\end{figure*}

\subsection{Main results}
\label{sec:main_results}
To validate our approach, we compare GCGSim against nine competitive baselines: SimGNN~\cite{b9}, GraphSim~\cite{b10}, GMN~\cite{b11}, MGMN~\cite{b12}, ERIC~\cite{b23}, NA-GSL~\cite{b25}, GEDGNN~\cite{b42}, GEDIOT~\cite{cheng2025computing}, and GraSP~\cite{zheng2025grasp}. 
We reimplement all methods, tuning their hyperparameters for optimal performance. 
To ensure a fair comparison, the output similarity scores of all models are unified to the definition used in this paper.

The results, summarized in Table~\ref{tab_1}, show that GCGSim consistently achieves state-of-the-art (SOTA) performance across the majority of metrics and datasets, demonstrating its effectiveness and robustness.
On AIDS700nef and LINUX, our model achieves the lowest Mean Squared Error (MSE), reducing the error by 13.8\% and 32.0\% respectively over the strongest competitor, indicating superior prediction accuracy. 
On IMDBMulti, GCGSim excels in ranking-based metrics, securing the top scores for $\rho$ (0.949) and $\tau$ (0.848), crucial for graph retrieval tasks.
Similarly, on PTC, it obtains the best performance on most metrics, including MSE and $\rho$, while some baselines face Out-of-Memory (OOM) issues.

In summary, the superior performance across diverse graph types demonstrates that learning from a GED-consistent perspective enables GCGSim to capture more accurate and robust similarity patterns than  previous approaches.

\subsection{Ablation study}
\label{sec:ablation}

We conducted an ablation study by individually removing four key components: GNCM, PSGD, IIR, and ECP. 
As presented in Table~\ref{tab_2}, removing any module degrades performance, confirming that all are integral and contribute synergistically to our framework.

The results reveal that PSGD is the cornerstone of our model. Its removal causes the most drastic performance drop, with the MSE on IMDBMulti more than doubling (from 0.568 to 1.196). 
This strongly validates that our theoretically-grounded disentanglement is fundamental to the model's success. 
Crucially, the removal of ECP also leads to a substantial performance degradation across all datasets (e.g., MSE on PTC increases by 25.8\%). This empirically validates our core motivation: that explicitly supervising the model with substructure-level edit costs is vital for learning a meaningful and accurate similarity function.
Similarly, removing GNCM results in a significant decline, confirming the necessity of generating pair-aware contextual representations.
Finally, the absence of IIR causes the mildest drop, demonstrating its role as a beneficial regularizer for learning canonical representations.
Collectively, these results verify our design choices and the contribution of each component.

\begin{table}[htbp]
\caption{Inference time comparison ($\times 10^{-5}$ s). Bold means the best result and underline means the second result.}
\label{tab_com}

\begin{center}
    \resizebox{0.85\linewidth}{!}{
    \begin{tabular}{cccc}
    \toprule
    \textbf{Method} & \textbf{AIDS700nef} & \textbf{LINUX} & \textbf{IMDBMulti} \\
    \midrule
    SimGNN & 12.23 & 59.49 & 60.7 \\
    GraphSim & 59.70 & 11.58 & 11.8 \\
    MGMN & 8.67 & 8.49 & 83.8 \\
    GEDGNN & 2.78 & 2.56 & 5.43 \\
    NA-GSL & 2.56 & 1.97 & 2.75 \\
    ERIC & \textbf{1.874} & \textbf{0.93} & \textbf{0.87} \\
    GEDIOT & 2.65 & 2.71 & 5.28 \\
    GraSP & 2.11 & 1.25 & 1.32 \\
    GCGSim  & \underline{1.986} & \underline{1.08} & \underline{1.19}\\
    \bottomrule
    \end{tabular}
    }
\end{center}
\end{table}

\subsection{Quantitative Analysis of Disentanglement via Mutual Information}
\label{sec:mi_experiment}

While the ablation study (Section~\ref{sec:ablation}) demonstrates the performance contribution of the PSGD module, we seek to more directly and quantitatively verify its core function: the disentanglement of aligned and unaligned representations. We hypothesize that if PSGD is effective, it should enforce informational orthogonality between the two representation types. In other words, the mutual information (MI) between the aligned embedding ($H_{as}$) and the unaligned embedding ($H_{us}$) should be minimized, signifying that they capture distinct and non-overlapping semantic features.

To measure this, we devise an experiment on the AIDS700nef and LINUX datasets. For a single anchor graph $G_i$ from the test set, we form a series of graph pairs by matching it with every other graph $\{G_j\}_{j \neq i}$ in the set. This process generates two corresponding sets of representations for the anchor graph $G_i$: a set of its aligned embeddings $\{H_{as_{i|j}}\}_{j \neq i}$ conditioned on each $G_j$, and a set of its unaligned embeddings $\{H_{us_{i|j}}\}_{j \neq i}$. These two sets effectively form the empirical distributions for the random variables $H_{as_i}$ and $H_{us_i}$. We then employ the Mutual Information Neural Estimator (MINE)~\cite{belghazi2018mutual}, a robust method for estimating MI in high-dimensional settings, to compute $I(H_{as_i} ; H_{us_i})$. This analysis is performed on the output of each of the four GNN layers for both our full GCGSim model and the GCGSim (w/o PSGD) variant to observe how disentanglement evolves with network depth.

The results, presented in Table~\ref{tab:mi_results}. 
Across both datasets and all network layers, the full GCGSim model exhibits substantially lower mutual information between its aligned and unaligned representations compared to the variant without PSGD. On AIDS700nef, the MI plummets by nearly 90\% at Layer 1 (from 3.46 to 0.37), and a similarly dramatic reduction of over 82\% is observed on LINUX (from 3.11 to 0.54).
This powerful initial separation is then effectively maintained throughout the subsequent layers. This substantial reduction in MI confirms that PSGD successfully learns semantically independent features for aligned and unaligned substructures. 


\begin{table}[t]
\caption{Mutual Information $I(H_{as} ; H_{us})$ on two datasets. Lower values indicate stronger disentanglement. 
}
\label{tab:mi_results}
\begin{center}
    \resizebox{0.85\linewidth}{!}{
    \begin{tabular}{c cc cc}
    
    \toprule
    \multirow{2}{*}{\textbf{\makecell{GNN\\Layer}}} & \multicolumn{2}{c}{\textbf{AIDS700nef}} & \multicolumn{2}{c}{\textbf{LINUX}} \\
    \cmidrule(lr){2-3} \cmidrule(lr){4-5}
    & w/o PSGD & GCGSim & w/o PSGD & GCGSim \\ \midrule
     1 & 3.46±0.41 & \textbf{0.37±0.12} & 3.11±0.06 & \textbf{0.54±0.35} \\
     2 & 2.11±0.36 & \textbf{0.47±0.27} & 2.89±0.12 & \textbf{0.58±0.08} \\
     3 & 2.49±0.67 & \textbf{0.47±0.15} & 3.00±0.10 & \textbf{0.75±0.13} \\
     4 & 2.98±0.51 & \textbf{0.73±0.25} & 2.57±0.14 & \textbf{0.48±0.24} \\ \bottomrule
    \end{tabular}%
    }
\end{center}
\end{table}

\begin{table*}[htbp]
\caption{Ablation study on the key components of GCGSim. Bold means the best result.}
\label{tab_2}
\begin{center}
    \resizebox{0.865\linewidth}{!}{
    \begin{tabular}{ccccccccccc}
    \toprule
    \multirow{2}{*}{\textbf{method}} & \multicolumn{5}{c}{\textbf{AIDS700nef}} & \multicolumn{5}{c}{\textbf{LINUX}} \\
    \cmidrule(lr){2-6} \cmidrule(lr){7-11}
    & MSE$\downarrow $ & $\rho \uparrow $ & $\tau \uparrow$ & $p@10 \uparrow $ & $p@20 \uparrow $ & MSE$\downarrow $ & $\rho \uparrow$ & $\tau \uparrow$ & $p@10 \uparrow$ & $p@20 \uparrow$ \\
    \midrule 
    GCGSim & \textbf{1.069} & \textbf{0.923} &\textbf{0.772} & \textbf{0.767} & \textbf{0.815} & \textbf{0.066} & \textbf{0.990} & \textbf{0.915} & \textbf{0.993} & \textbf{0.997} \\
    GCGSim(w/o GNCM) & 1.221 & 0.914 & 0.760 & 0.733 & 0.794 & 0.094 & 0.988 & 0.913 & 0.981 & 0.991 \\
    GCGSim(w/o PSGD)& 1.353 & 0.908 & 0.751 & 0.727 & 0.776 & 0.113 & 0.969 & 0.912 & 0.970 & 0.973 \\
    GCGSim(w/o IIR)& 1.094 & 0.918 & 0.760 & 0.759 & 0.796 & 0.088 & 0.987 & 0.913 & 0.982 & 0.989 \\
    GCGSim(w/o ECP)& 1.285 & 0.892 & 0.731 & 0.702 & 0.756 & 0.099 & 0.975 & 0.898 & 0.963 & 0.978 \\
    \bottomrule
    \end{tabular}}
\end{center}

\begin{center}
    \resizebox{0.865\linewidth}{!}{
    \begin{tabular}{ccccccccccc}
    \toprule
    \multirow{2}{*}{\textbf{method}} & \multicolumn{5}{c}{\textbf{IMDBMulti}} & \multicolumn{5}{c}{\textbf{PTC}} \\
    \cmidrule(lr){2-6} \cmidrule(lr){7-11} 
    & MSE$\downarrow $ & $\rho \uparrow $ & $\tau \uparrow$ & $p@10 \uparrow $ & $p@20 \uparrow $ & MSE$\downarrow $ & $\rho \uparrow$ & $\tau \uparrow$ & $p@10 \uparrow$ & $p@20 \uparrow$ \\
    \midrule
    GCGSim & \textbf{0.568} & \textbf{0.949} & \textbf{0.848} & \textbf{0.892} & \textbf{0.905} & \textbf{1.548} & \textbf{0.940} & \textbf{0.806} & \textbf{0.617} & \textbf{0.722} \\
    GCGSim(w/o GNCM) & 0.775 & 0.897 & 0.832 & 0.858 & 0.876 & 1.662 & 0.934 & 0.798 & 0.604 & 0.684 \\
    GCGSim(w/o PSGD) & 1.196 & 0.925 & 0.846 & 0.880 & 0.896 & 2.055 & 0.936 & 0.800 & 0.574 & 0.666 \\
    GCGSim(w/o IIR) & 0.608 & 0.935 & 0.840 & 0.878 & 0.889 & 1.865 & 0.935 & 0.799 & 0.569 & 0.687 \\
    GCGSim(w/o ECP) & 0.796 & 0.897 & 0.803 & 0.831 & 0.842 & 1.947 & 0.916 & 0.775 & 0.558 & 0.643 \\
    \bottomrule
    \end{tabular}}
\end{center}
\end{table*}

\subsection{Verifying the Semantics of Disentangled Representations}
\label{sec:feasibility_study}

A core objective of our work is to learn disentangled representations, $H_{as}$ and $H_{us}$, that meaningfully correspond to aligned and unaligned substructures. 
However, the absence of ground-truth substructure labels makes direct validation challenging.
To circumvent this, we design a set of \textit{embedding swapping} experiments to probe the semantic properties of the learned representations. 
Our central hypothesis is that if the disentanglement is successful, these embeddings should exhibit specific, predictable behaviors when swapped within or across graph pairs.

We introduce two probing tasks: Intra-Instance Swap (IIS) and Extra-Instance Swap (EIS). We use the Mean Squared Error (MSE) difference relative to the standard GCGSim model (without swapping) as the primary evaluation metric. A larger MSE difference indicates a greater performance degradation caused by the swap.

\begin{itemize}
    \item \textit{Intra-Instance Swap (IIS):} For a given graph pair $(G_i, G_j)$, we swap its learned aligned embeddings, i.e., we feed $(H_{as_j}, H_{as_i})$ instead of $(H_{as_i}, H_{as_j})$ into the downstream prediction head during inference. 
    \textit{Rationale:} If our model, particularly through the IIR mechanism, has learned a canonical and interchangeable representation for "aligned-ness", this swap should have a minimal impact on performance. A significant performance drop would imply the embeddings are source-graph-specific.

    \item \textit{Extra-Instance Swap (EIS):} This experiment tests a stronger condition of context-invariance. We create two distinct graph pairs, $(G_i, G_j)_1$ and $(G_k, G_l)_2$, that are synthetically constructed to share identical substructures of a specific type.
    \begin{itemize}
        \item \textit{EISA (Aligned Swap):} The pairs are constructed such that their aligned substructures are identical. We then swap the aligned embeddings between these pairs, i.e., using $(H_{as_i}, H_{as_j})_2$ to predict the similarity for pair 1. 
        \textit{Rationale:} This directly tests if $H_{as}$ encodes features of the aligned structure itself, independent of the surrounding unaligned context from its original host graph.
        
        \item \textit{EISU (Unaligned Swap):} Symmetrically, the pairs are constructed to share identical unaligned substructures. We then swap the unaligned embeddings $(H_{us_i}, H_{us_j})$ between them. 
        \textit{Rationale:} This probes the semantic purity and context-invariance of the unaligned representations.
    \end{itemize}
\end{itemize}

The results of these experiments are presented in Fig.~\ref{fig:swap_experiments}.
Our analysis aims to rule out potential trivial solutions and validate that a meaningful disentanglement has been achieved.

A successful disentanglement must avoid two trivial solutions: (1)~$H_{as}$ is a vacuous representation, and all predictive information is stored in $H_{us}$; (2)~$H_{us}$ is vacuous, and all information is in $H_{as}$.

The results from EISA (Fig.~\ref{fig:swap_experiments}b, 'All' model) 
robustly refute the first trivial solution. Swapping aligned embeddings from a completely independent graph pair causes only a minor increase in MSE. If $H_{as}$ were vacuous, the model would be relying on the now-mismatched unaligned embeddings ($H_{us}$ from pair 1 with $H_{us}$ from pair 2), which would have led to a catastrophic performance collapse. The observed stability strongly implies that $H_{as}$ captures context-invariant, semantically meaningful features of the aligned substructures. Symmetrically, the EISU results (Fig.~\ref{fig:swap_experiments}c) 
challenge the second trivial solution. The moderate MSE increase indicates that $H_{us}$ also captures essential predictive information.

The IIS results (Fig.~\ref{fig:swap_experiments}a) highlight the effectiveness of our full model. While swapping aligned embeddings significantly degrades the performance of a model without our proposed components ('None'), the complete GCGSim ('All') is remarkably robust to this swap. This demonstrates that our framework 
successfully enforces the learning of canonical and interchangeable aligned representations.

Collectively, these experiments provide compelling evidence against trivial solutions and affirm that GCGSim learns to disentangle representations in a semantically meaningful way. The model successfully separates features corresponding to aligned and unaligned substructures into their respective representations. While the moderate performance drop in EISU suggests that learning perfectly clean, context-free unaligned representations remains challenging, our overall framework demonstrably achieves its goal of structured and semantically valid representation disentanglement, providing a solid foundation for its superior performance.

\begin{figure}[t]
  \centering
  \includegraphics[width=0.90\linewidth]{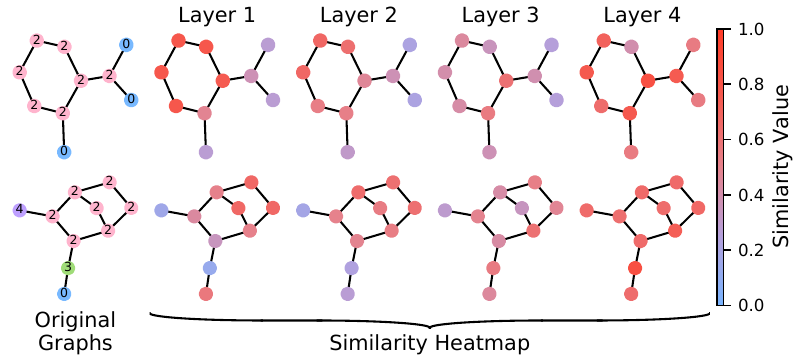} 
  \caption{Visualization of the pair-aware similarity learned by the GNCM module across GNN layers. The color intensity indicates the similarity of a node's evolving local context to the other graph's global structure (blue: low, red: high).}
  \label{fig:gncmmap}
\end{figure}

\subsection{Visual Analysis of the GNCM Module}
\label{sec:qualitative_gncm}

To gain deeper insight into the inner workings of our Graph-Node Cross Matching (GNCM) module, we visualize the learned similarities between the local representations of one graph and the global representation of the other. As illustrated in Fig.~\ref{fig:gncmmap}, we generate a heatmap of these similarity scores for each node across all GNN layers.

A primary observation from the heatmaps is that the GNCM module successfully learns to differentiate between nodes belonging to aligned versus unaligned substructures. Across all layers, nodes within aligned substructures consistently exhibit higher similarity scores (indicated by red and pink hues). Conversely, nodes from unaligned parts of the graph, such as the distinct node within the annular substructure in Layer 3, show markedly lower similarity values (cooler, blueish colors). 

The visualization also reveals a compelling dynamic of how the basis for similarity assessment evolves with network depth, a phenomenon directly tied to the expanding receptive field of the GNN.
\begin{itemize}
    \item \textit{In shallow layers (e.g., Layers 1-2)}, where a node's representation is dominated by its immediate neighbors and its own features, the similarity scores are heavily influenced by intrinsic node attributes (e.g., labels). The model primarily performs local, feature-based matching.
    \item \textit{In deeper layers (e.g., Layers 3)}, as the receptive field grows to encompass larger topological patterns, the structural context begins to dominate the similarity computation. The model shifts from local feature matching to a more holistic, topology-aware comparison.
\end{itemize}
Interestingly, in the final layer, the similarities tend to become more uniform across all nodes. This is a well-known characteristic of deep GNNs, where node representations can converge due to over-smoothing. This observation highlights the importance of using multi-layer information and reinforces the motivation for our approach, which does not rely solely on the output of the final layer for its decision.

\begin{figure}[htbp]
	\centering
	\begin{minipage}{0.48\linewidth}
        \subfloat[$\lambda$]{\includegraphics[width=0.98\linewidth]{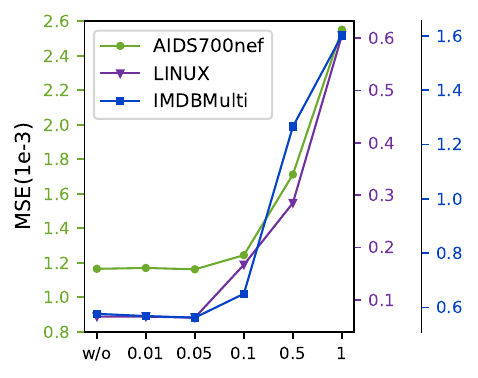}}
	\end{minipage}
	\begin{minipage}{0.48\linewidth}
        \subfloat[$\beta $]{\includegraphics[width=0.98\linewidth]{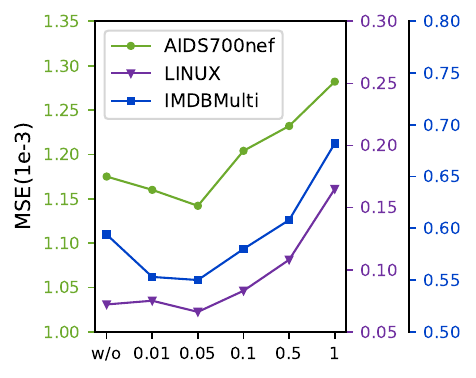}}
	\end{minipage}
        \caption{Sensitivity analysis on the ECP weight $\lambda$ (a) and the IIR parameter $\beta$ (b).}
    \label{fig:param_sensitivity}
\end{figure}

\subsection{Complexity analysis}

The time complexity associated with the generation of node embeddings and graph embeddings is $O\left( \left| E \right| \right) $, where $\left| E \right|$ is the number of edges of the graph, in the Embedding Learning stage. The GNCM has complexity $O \left(max\left(N_i,N_j\right)d’\right))$, which $d'$ is the dimension of node embeddings and graph embeddings. The time complexity for Aligned Discriminator and Unaligned Discriminator is $O\bigl({d’}^2K\bigr)$, $K$ is the feature map dimension of the NTN. Thus the complexity of GCGSim is $O\bigl( \left| E \right| + \left(max\left(N_i,N_j\right)d’\right) + {d’}^2K \bigr)$.

\begin{figure}[htbp]
\centering
\subfloat[AIDS700nef]{\includegraphics[width=0.95\linewidth]{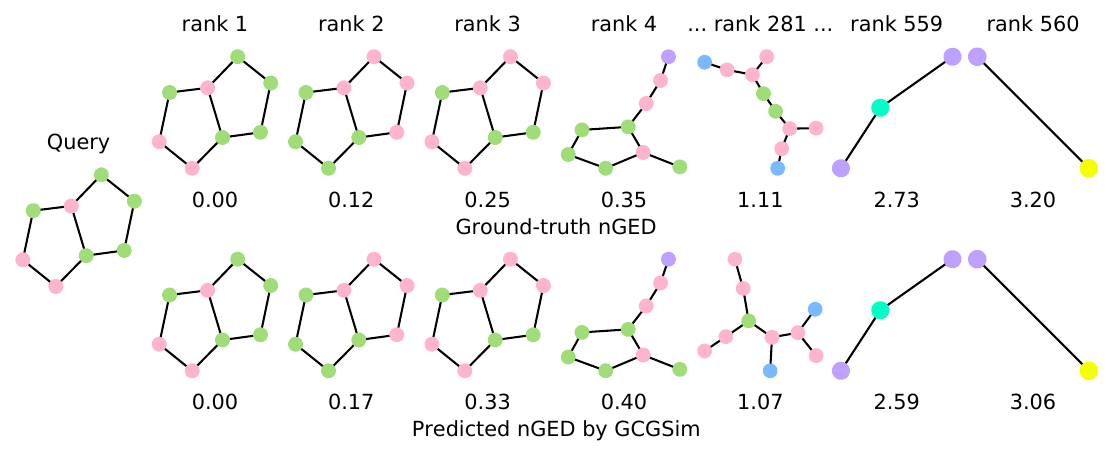}
\label{fig_6_1}}
\hfil
\subfloat[LINUX]{\includegraphics[width=0.95\linewidth]{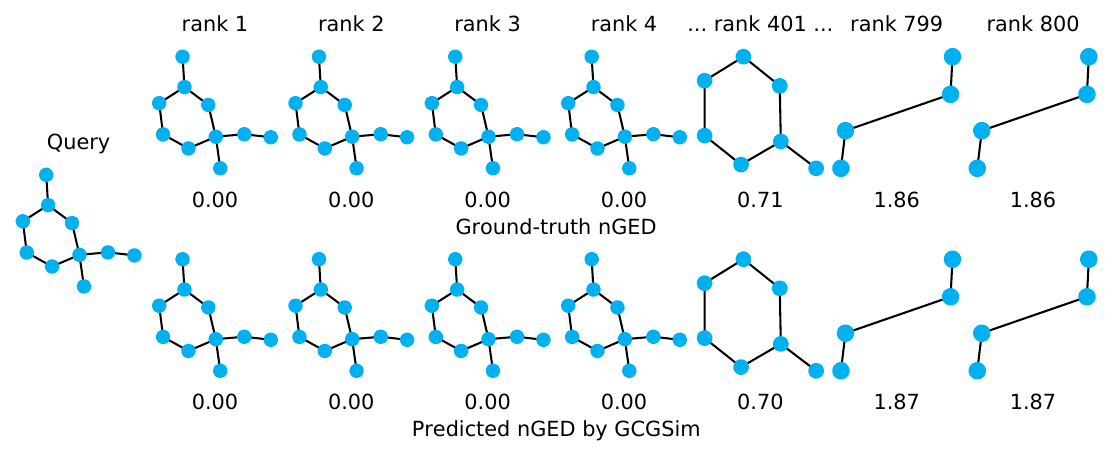}
\label{fig_6_2}}
\hfil
\subfloat[IMDBMulti]{\includegraphics[width=0.95\linewidth]{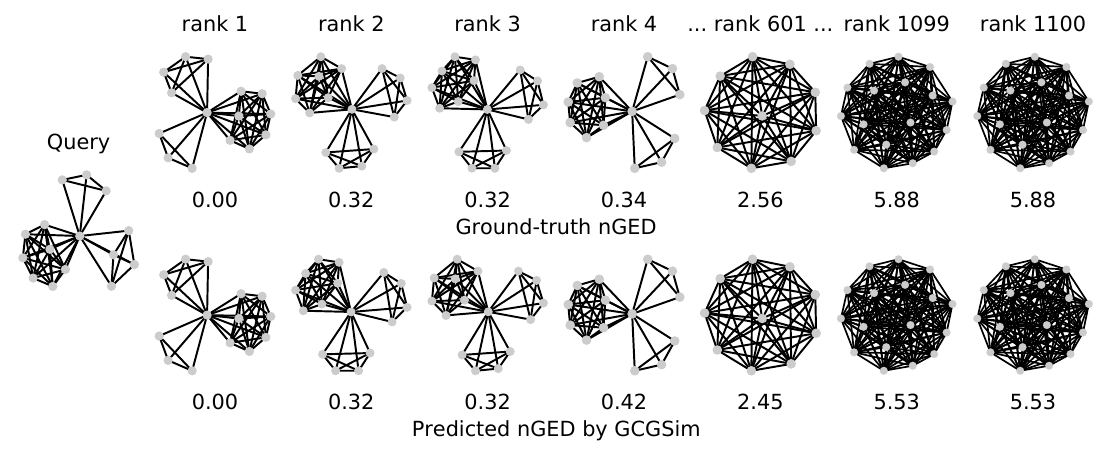}
\label{fig_6_3}}
\caption{A sample of ranking results under graph similarity score. Nodes with different labels are assigned different colors.
}
\label{fig_6}
\end{figure}

\subsection{Computational Efficiency}
\label{sec:efficiency}

To assess the practical applicability of our model, we compare its inference time against the baselines, with results presented in Table~\ref{tab_com}. 
The comparison reveals that GCGSim is highly efficient, consistently ranking as the second-fastest method across all datasets and closely following the state-of-the-art model ERIC.

Notably, GCGSim is significantly faster than many established methods like GraphSim and MGMN, which often employ more computationally intensive node-level interaction modules.
The marginal difference in speed compared to ERIC is a direct trade-off for our more principled architecture, which includes disentanglement encoders and an auxiliary prediction task. 
This minor computational overhead is justifiable, as it enables the superior predictive accuracy demonstrated in Section~\ref{sec:main_results}.
In conclusion, GCGSim strikes an excellent balance between state-of-the-art accuracy and high computational efficiency, making it a practical solution for real-world applications. 

\subsection{Parameter sensitivity}
We analyze the sensitivity of the ECP weight $\lambda$ and the IIR parameter $\beta$, with results shown in Fig~\ref{fig:param_sensitivity}. 
Both hyperparameters exhibit a similar U-shaped impact on performance. 
As observed, the model achieves optimal performance across all datasets when both $\lambda$ and $\beta$ are set to approximately 0.05. 
Performance degrades when the components are removed ('w/o' case), consistent with our ablation study. 
Conversely, performance also deteriorates sharply with excessively large values. 
This is because an overly large $\lambda$ compromises the primary similarity task, while a large $\beta$ can introduce disruptive noise to the representations. 
Therefore, we set $\lambda=0.05$ and $\beta=0.05$ in our main experiments.

\subsection{Case study}
We conduct graph search experiments on AIDS700nef, LINUX, and IMDBMulti to retrieve k graphs from the dataset that are most similar to the given query graph. As shown in Fig.~\ref{fig_6}, in each example, we demonstrate the predicted similarity ranking result computed by GCGSim compared with ground-truth ranking. The top-ranked graph has a high degree of isomorphism with the query. It indicates that GCGSim possesses the capability to retrieve graphs that are similar to the query graph.

\section{Conclusion}
In this paper, we addressed the fundamental mismatch between prevalent GNN-based methods and the core principles of GED. We proposed GCGSim
that reformulates 
the GSC task 
from a GED-consistent perspective of graph-level matching and substructure-level edit costs.
This is achieved through a synergy of three components designed to tackle distinct conceptual challenges: GNCM, to overcome the context-agnostic limitations of conventional graph embeddings; the theoretically-grounded PSGD, to operationalize the core GED concept of partitioning graphs into aligned (similar) and unaligned (dissimilar) substructures within the representation space; and IIR,to enforce canonicity upon the representations of aligned substructures. Extensive experiments show GCGSim achieves state-of-the-art performance, confirming that embedding these combinatorial principles directly into the learning objective yields a more effective and robust model for graph similarity.

\bibliographystyle{IEEEtran}
\bibliography{sample-base}

\end{document}